\documentclass[journal]{IEEEtran}
\usepackage{multirow}
\usepackage{makecell}
\usepackage{graphicx}
\usepackage{amssymb}

\ifCLASSINFOpdf
\else
\fi
\begin{document}
%
\title{PainSeeker: An Automated Method for Assessing Pain in Rats Through Facial Expressions}
%
%
%

\author{Liu~Liu$^\dag$,
Guang~Li$^\dag$,
Dingfan~Deng,
Jinhua~Yu$^*$,
and~Yuan~Zong$^*$,~\IEEEmembership{Member,~IEEE}
\thanks{The animal experiments in this work have been approved by the Institutional Animal Use Committee of NMU with the Approval No. IACUC-2305045.}
\thanks{L. Liu and J. Yu are with the Affiliated Stomatological Hospital, Nanjing Medical University, Nanjing 210029, China. (email: yujinhua@njmu.edu.cn)}
\thanks{G. Li, D. Deng, and Y. Zong are with the School of Biological Science and Medical Engineering, Southeast University, Nanjing 210096, China. (email: xhzongyuan@seu.edu.cn)}
\thanks{$^\dag$ Equal contributions. $^*$ Corresponding authors.}
}

%

%

\markboth{Journal of \LaTeX\ Class Files,~Vol.~14, No.~8, August~2015}%
{Shell \MakeLowercase{\textit{et al.}}: Bare Demo of IEEEtran.cls for IEEE Journals}
%




\maketitle

\begin{abstract}
In this letter, we aim to investigate whether laboratory rats' pain can be automatically assessed through their facial expressions. To this end, we began by presenting a publicly available dataset called RatsPain, consisting of 1,138 facial images captured from six rats that underwent an orthodontic treatment. Each rat' facial images in RatsPain were carefully selected from videos recorded either before or after the operation and well labeled by eight annotators according to the Rat Grimace Scale (RGS). We then proposed a novel deep learning method called PainSeeker for automatically assessing pain in rats via facial expressions. PainSeeker aims to seek pain-related facial local regions that facilitate learning both pain discriminative and head pose robust features from facial expression images. To evaluate the PainSeeker, we conducted extensive experiments on the RatsPain dataset. The results demonstrate the feasibility of assessing rats' pain from their facial expressions and also verify the effectiveness of the proposed PainSeeker in addressing this emerging but intriguing problem. The RasPain dataset can be freely obtained from https://github.com/xhzongyuan/RatsPain.
\end{abstract}


%
\IEEEpeerreviewmaketitle

\section{Introduction}
%
%
%
%
\IEEEPARstart{R}{ats} are one of the most extensively used laboratory animals, which have significantly contributed to the advancement of research in biology and medicine~\cite{institute1986guide,rees2005animal}. In numerous experiments involving rats, such as the animal testing of painkillers, the assessment of their pain levels is a critical and indispensable step~\cite{feng2019new}. However, it should be noted that assessing pain in rats is not an easy task. Unlike humans, rats lack the ability to express their feelings through language, making it impossible for them to provide direct feedback on their pain levels~\cite{broome2023going}. Despite this difficulty, researchers have developed lots of effective methods over the past few decades~\cite{turner2019review,sadler2022innovations}. 

Behavioral testing methods are currently the predominant approaches for assessing pain in rats. These methods can be roughly classified into two types: evoked behavioral testing and non-evoked behavioral testing methods~\cite{sadler2022innovations}. The Von Frey test is a well-known evoked method~\cite{chaplan1994quantitative,pitcher1999paw}. This test involves applying a set of calibrated filaments with progressively increasing bending forces to a targeted area of a rat, such as the paw, until the filament slightly bows against the skin. Observing this slight bow allows us to obtain the rat's pain level determined by its corresponding force threshold. Although the Von Frey test is a commonly used and relatively quick method for assessing pain in rats, it is crucial to acknowledge that this method relies on the animals' behavioral reactions to repeated invasive stimuli, which raises concerns regarding laboratory animal welfare. As for the non-evoked methods, it is worth mentioning wheel running analysis, which offers a non-invasive approach to assess pain in rats~\cite{grace2014suppression,kandasamy2016home}. In this method, the rat's wheel running activities are continuously monitored at regular intervals. Decreased activities of rats, e.g., running distance and time spent on the wheel, may indicate the presence of pain in rats. However, the wheel running analysis, like many other spontaneous behavior testing methods, is time-consuming. It requires at least one week of training for rats as preparatory work such that the rats can run on the wheel~\cite{legerlotz2008voluntary}. This creates a barrier to promptly assess the rats' pain.



In recent years, researchers have tried to investigate whether pain in rats can be accurately and effectively assessed in a non-invasive manner without a long period. Motivated by the exploration of the relationship between facial action units and pain expressions for humans in fields of computer vision and psychology~\cite{ekman1997face,rojo2015pain}, Sotocinal et al.~\cite{sotocinal2011rat} developed a pain assessment tool for rats called the Rat Grimace Scale (RGS). This tool quantifies spontaneous pain in rats through blinded coding of their facial expressions and has been demonstrated the effectiveness and convenience. In contrast to existing behavioral testing methods, the RGS allows us to conveniently and quickly estimate pain levels in rats based solely on their facial expression images.


\begin{figure*}[t!]
\centering
\includegraphics[width=\textwidth]{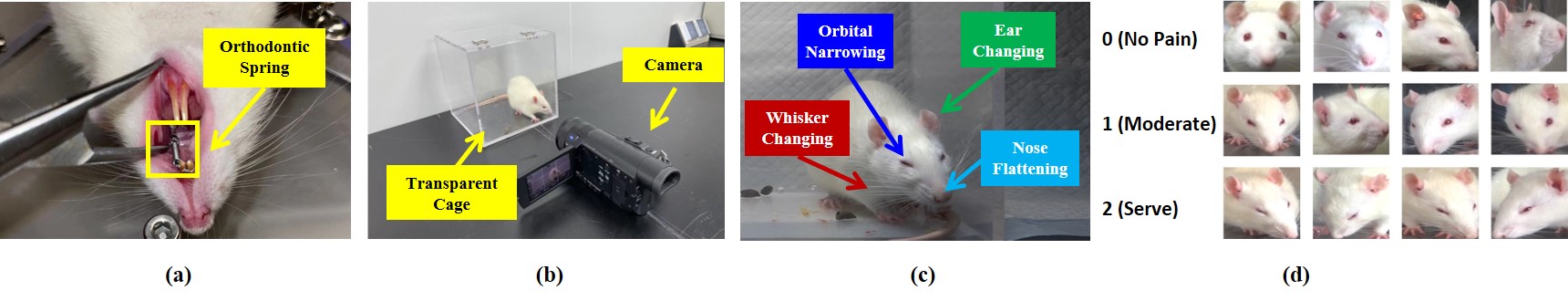}
\caption{Illustration of Rats' Facial Image Collection and Pain Annotation: (a) Rat Undergoing Orthodontic Treatment, (b) Setup of the Shooting Environment and Equipment, and (c) Diagram Depicting the Labeling of Pain Levels in Rats Using the Rat Grimace Scale (RGS)~\cite{sotocinal2011rat}. (d) Samples of Facial Images Expressing Varying Pain Levels in Rats from the RatsPain Dataset.}
\label{fig:nb1}
\end{figure*}

Motivated by the work of RGS and recent advancements in automated facial pain assessment for other laboratory animals and humans~\cite{broome2023going,gkikas2023automatic}, we try to develop automated methods for assessing pain in rats through facial expressions. Specifically, we first establish a platform called RatsPain dataset for evaluating automated methods by collecting 1,138 well-labeled facial images from six rats that underwent an orthodontic treatment. Second, we propose a novel deep learning method called PainSeeker for seeking pain-related and pose-invariant facial local regions of rats to learn discriminative features from the facial images. The experimental results on the collected RatsPain dataset demonstrated the effectiveness of the proposed PainSeeker method, thus supporting the feasibility of assessing pain in rats through their facial expressions.

\section{RatsPain Dataset}

In this section, we present the procedure for collecting and labeling facial images of rats related to pain to construct the RatsPain dataset, as depicted in Fig.~\ref{fig:nb1}. Initially, orthodontic treatment was conducted on six healthy male Sprague-Dawley (SD) rats, each aged about eight weeks, to induce pain, as illustrated in Fig.~\ref{fig:nb1}(a). The orthodontic spring used in the treatment applied a force of $0.8N$. It is important to note that each rat was housed in an individual transparent cage, equipped with a front-facing camera, as shown in Fig.~\ref{fig:nb1}(b). The camera recorded one-hour videos of each rat before and after the orthodontic treatment, respectively. Notably, since pain in rats generally peaks within about three days after orthodontic treatment~\cite{horinuki2015orthodontic}, the postoperative videos were obtained twenty-four hours after the treatment, while the preoperative videos were captured one hour before the treatment.

\subsection{Selection of Rat's Facial Images}

Once the recording was completed, we established strict inclusion and exclusion criteria to guide the subsequent manual selection of high-quality rats' facial images related to pain. Specifically, an eligible facial image of a rat should clearly display the four key facial components associated with pain in rats~\cite{sotocinal2011rat}: the eye, ear, whisker, and nose, as shown in Fig.~\ref{fig:nb1}(c). To ensure the sample diversity, at most five images were uniformly extracted from the video per minute. It is also worth noting that partial obstruction of one side of the eye, whisker, or both is acceptable, as their absence does not significantly affect the assessment of pain based on the corresponding facial images. Additionally, since images depicting rats in sleeping, standing, or grooming states cannot accurately assess their pain levels through facial expressions, we excluded them from the selection process. To carry out the selection process, we enlisted seven undergraduate/graduate students with biomedical engineering or medicine backgrounds. Five students were responsible for capturing the images, while the remaining two reviewed and verified whether the captured images met the criteria. As a result, we obtained 1,295 high-quality facial images of rats specifically related to pain.

\subsection{Annotation of Pain Levels in Rats}

To obtain high-confidence image-level pain labels for the selected facial images of rats mentioned above, we recruited eight additional well-trained undergraduate/graduate students. These students implemented a two-stage labeling scheme based on the RGS~\cite{sotocinal2011rat}. In the first stage, five students were responsible for assigning one of three pain scores to each image based on four pain-related facial components, shown in Fig.~\ref{fig:nb1}(c). According to the RGS, these pain scores were defined as 0 (no pain), 1 (moderate pain), and 2 (severe pain), respectively. However, some students faced challenges in determining the appropriate pain score for certain components, such as the whisker, due to image quality issues. To address this, we allowed for an "uncertain" component-level pain label to account for these cases.


\begin{table}[t!]
\centering
\footnotesize
\renewcommand{\arraystretch}{1.2}
\caption{The Sample Statistical Information of the RatsPain Dataset.}
\begin{tabular}{|c|c|c|c|}
\hline
\textbf{Sample Group} & \textbf{Pain Label} & \multicolumn{2}{c|}{\textbf{\# Facial Images}} \\ \hline\hline
\multirow{3}{*}{High Confidence} & No & ~~~541~~~ & \multirow{3}{*}{~~1,138~~}\\ \cline{2-3}
& Moderate & 591& \\ \cline{2-3}
& Severe & 6 & \\ \hline
Low Confidence & N/A & \multicolumn{2}{c|}{157} \\ \hline\hline
\multicolumn{2}{|c|}{Total} & \multicolumn{2}{c|}{1,295}\\\hline
\end{tabular}
\label{tab:nb1}
\end{table}

In the second stage, the remaining three students continued to provide their scores for the facial images containing components with low confidence. Here, low confidence is defined as less than four students assigning the same score in the first stage. We then examined the score distribution of the total eight pain scores for all the components with low confidence. Only components that received at least five consistent scores were accepted. Using this two-stage labeling scheme, we obtained three or four highly confident component-level pain scores for 1,138 out of 1,295 images. The rounded average score value was used as the image-level pain label for each facial image of the rats. However, for the remaining 157 images, their image-level pain labels were not assigned due to a significant number of components with low confidence. To provide readers with a glance at our RatsPain dataset, we have included its statistical information in Table~\ref{tab:nb1} and sample facial images expressing different levels of pain in Fig.~\ref{fig:nb1}(d).

\begin{figure}[t!]
\centering
\includegraphics[width=\columnwidth]{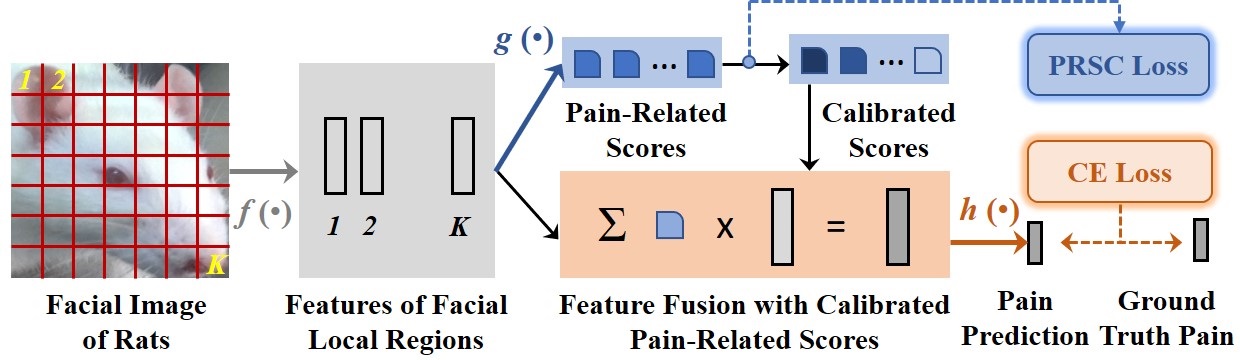}
\caption{Overall Structure of PainSeeker for Assessing Pain in Rats Through Facial Expressions.}
\label{fig:nb3}
\end{figure}

\section{PainSeeker for Pain Assessment in Rats}

\subsection{Basic Idea and Preliminary}

The key concept behind PainSeeker emerged from our observations while housing rats in transparent cages. We noticed that due to their lively nature and discomfort caused by orthodontic treatment, rats frequently exhibit erratic movements. These movements result in frequent changes in head pose and non-frontal perspectives in facial images captured by a fixed-position camera. This poses a challenge in learning discriminative features from their facial expressions to assess pain. To address this issue, we propose a novel deep learning model called PainSeeker, as shown in Fig.~\ref{fig:nb3}. The objective of PainSeeker is to seek pain-related facial local regions that are robust to head pose invariance, allowing for the learning of pose-invariant and pain-discriminative features from facial images of rats.

In what follows, we will provide a detailed description of the PainSeeker. We begin by introducing a training set, which comprises $N$ facial images of rats denoted as $\{\mathcal{X}_1,\cdots,\mathcal{X}_N\}$. Each image, $\mathcal{X}_i \in \mathbb{R}^{d\times d\times 3}$, represents a color image with three channels and dimensions $d\times d$. The corresponding label is denoted as $\mathbf{y}_i \in \mathbb{R}^{c\times 1}$, which is a one-hot vector generated based on the ground truth pain levels ranging from 1 to $c$. As depicted in Fig.~\ref{fig:nb3}, a facial image of a rat is first passed through a set of convolutional layers, resulting in a set of raw features corresponding to $K = M \times M$ facial local regions. These features can be represented as $[\mathbf{x}_{i,1},\cdots,\mathbf{x}_{i,K}] = reshape(f(\mathcal{X}_i),[K, d_x])$, where $reshape(\cdot)$ denotes the process of reshaping the tensor $f(\mathcal{X}_i)\in\mathbb{R}^{M\times M\times d_x}$ into a $d_x$-by-$K$ matrix. Here, $f(\cdot)$ represents the operation performed by the convolutional layers, $K$ represents the size of feature maps in the last convolutional layer, and $d_x$ denotes the number of feature maps.

\subsection{Seeking and Calibrating the Contributions of Pain-Related Facial Local Regions}

Instead of directly flattening these facial local region-related features like widely-used structure of convolutional neural network (CNN), e.g., VGG~\cite{simonyan2014very}, our PainSeeker model incorporates the attention mechanism~\cite{niu2021review} to assign a fully connected (FC) layer that measures the pain-related scores of features corresponding to each facial local region. This can be expressed as:
\begin{eqnarray}
\beta_{i,j} = \frac{\sigma(g(\mathbf{x}_{i,j}))}{\sum_{j=1}^K \sigma(g(\mathbf{x}_{i,j}))},~j=\{1,\cdots,K\},
\end{eqnarray}
where $g(\cdot)$ denotes the operation performed by the FC layer, and $\sigma(\cdot)$ represents the sigmoid function.

Using these pain-related scores, we can predict the pain level of a given facial image of rats by using the weighted fused feature, which can be expressed as: $\mathbf{y}_i^{p} = \textup{Softmax}(h(\sum_{j=1}^{K} \beta_{i,j} \mathbf{x}_{i,j}))$, where $h(\cdot)$ and $\textup{softmax}(\cdot)$ represent the operations performed by a FC layer and softmax function. To train the model, we employ the cross-entropy (CE) loss to establish the relationship between the pain label predicted by PainSeeker and the corresponding ground truth. The CE Loss can be written as:
\begin{eqnarray}
\mathcal{L}_{CE} = \mathcal{J}(\mathbf{y}_i,\mathbf{y}_i^{p}),
\label{eqn:nb3}
\end{eqnarray}
where $\mathcal{J}(\cdot)$ is the CE function.

It is important to note that minimizing the CE loss in Eq.(\ref{eqn:nb3}) allows us to obtain a set of $\beta_{i,j}$ values ranging from 0 to 1. These values help seek facial local regions that contribute to learning discriminative features for predicting the pain level of rats. However, due to the deviation of rats' head pose from a frontal perspective, the association between these contributive regions and pain is often reduced. Consequently, the values of these regions, which should ideally be larger, may exhibit a small gap compared to the less contributive regions. To address this issue, we propose a novel regularization term called pain-related score calibration (PRSC), which is derived from the widely-used triplet loss~\cite{weinberger2009distance}. The PRSC loss function is formulated as follows:
\begin{eqnarray}
\mathcal{L}_{PRSC} = \frac{1}{K_{h}} \sum_{j=1}^{K_{h}} \max\{0, \delta - (\beta_{i,j}^h - \bar\beta_{i}^{r})\},
\label{eqn:nb4}
\end{eqnarray}


In Eq.(\ref{eqn:nb4}), $\delta$ represents the preset margin value, $\beta_{i,j}^h$ denotes the $j^{th}$ element among the highest $K_{h}$ highly pain-related scores, and $\bar\beta_{i}^{r}$ represents the mean value among the $K_{r} = K - K_h$ resting scores corresponding to less contributive facial local regions. Note that minimizing the PRSC loss results in a large value gap between the highly pain-related scores and the remaining scores. This emphasizes the features learned from the highly pain-related facial local regions while suppressing the less contributive features in the original fused features. As a result, the fused features used for predicting pain levels become more pain-discriminative and robust to head pose variations.

\subsection{Total Loss Function of PainSeeker}

The total loss function of the proposed PainSeeker model is obtained by combining the loss functions presented in Eqs.(\ref{eqn:nb3}) and~(\ref{eqn:nb4}), and computing the summation over all $N$ training samples. The resulting expression is given as follows:
\begin{eqnarray}
\min_{\Theta} \frac{1}{N}\sum_{i=1}^{N} [\mathcal{J}(\mathbf{y}_i,\mathbf{y}_i^{p}) +  \frac{\lambda}{K_h}\sum_{j=1}^{K_{h}} \max\{0, \delta - (\beta_{i,j}^h - \bar\beta_{i}^{r})\}],
\label{eqn:nb4}
\end{eqnarray}
where $\Theta = \{\theta_f,\theta_g,\theta_h\}$ represent the parameters associated with the operations performed by the layers $f$, $g$, and $h$ in PainSeeker and $\lambda$ is the trade-off parameter to control the balance between the CE and PRSC losses.

\begin{table*}[t!]
\centering
\renewcommand{\arraystretch}{1.2}
\caption{Comparison of F1-Score and Accuracy for Pain Assessment in Rats Through Facial Expressions Under the LORO Protocol. Best Result is Highlighted in Bold.}
\begin{tabular}{|l|cccccc|c|}
\hline
\textbf{Method} & \textbf{Rat\#1} & \textbf{Rat\#2} & \textbf{Rat\#3} & \textbf{Rat\#4} & \textbf{Rat\#5} & \textbf{Rat\#6} & \textbf{LORO}\\\hline\hline
LBP$_{R1P8}~(4\times4)$ &  0.7862 / 71.95 &0.7292 / 64.29 & 0.5399 / 62.12& 0.6291 / 65.65 & 0.6923 / 64.84 & 0.6087 / 62.89 & 0.6854 / 65.64\\\hline
LBP$_{R1P8}~(8\times8)$ & 0.7835 / 71.49 & 0.7396 / 67.14 & 0.4906 / 59.09	& 0.6184 / 65.65 & 0.6300 / 59.34 & 0.5493 / 57.73 & 0.6853 / 64.24 \\\hline
LBP$_{R1P8}~(16\times16)$ & 0.7703 / 69.23& 0.7068 / 65.24 & 0.4648 / 61.62 & 0.5326 / 62.61 & 0.5930 / 55.49 & 0.7000 / 69.07 & 0.6462 / 63.62 \\\hline
LBP$_{R3P8}~(4\times4)$ & 0.8014 / 74.66 & 0.7341 / 66.19 & 0.5223 / 62.12	& 0.5842 /	63.48 & 0.7184 / 68.13	& 0.7736 / 75.26	& 0.6984 / 67.66 \\\hline
LBP$_{R3P8}~(8\times8)$ & 0.7286 / 66.97 & 0.7315 / 67.14 & 0.4247 / 57.58 & 0.6193 / 67.39 & 0.7264 / 68.13 & 0.6800 / 67.01 & 0.6689 / 65.64 \\\hline
LBP$_{R3P8}~(16\times16)$ & 0.7416 / 68.78 & 0.7160 / 65.24 & 0.4460 / 61.11 & 0.5510 /  61.74 & 0.5333 / 50.00 & 0.6346 / 60.82 & 0.6235 / 61.69 \\\hline
ResNet-18 & \textbf{0.8921} / \textbf{83.26} & 0.7387 / 72.38 & 0.5556 / 71.72 & 0.6098 / 58.26 & 0.8042 / 69.23 & 0.7914 / 70.10& 0.7562 / 70.83 \\\hline\hline
PainSeeker w/o PRSC  & 0.8548 / 75.57 & 0.7258 / 67.62 & 0.5472 / \textbf{75.76} & 0.6367 / 61.30 & 0.7874 / 70.33 & 0.8197 / 77.32 & 0.7513 / 70.56 \\\hline
PainSeeker & 0.8780 / 81.90 & \textbf{0.7778 / 75.24} & \textbf{0.5714} / 66.67 & \textbf{0.6446 / 62.61} & \textbf{0.8365 / 76.37} & \textbf{0.8750 / 83.51} & \textbf{0.7754 / 73.37}\\\hline
\end{tabular}
\label{tab:nb2}
\end{table*}

\section{Experiments}

\subsection{Experimental Protocol and Implementation Detail}

In this section, we conduct extensive pain assessment experiments on the high-confidence sample set of our RatsPain dataset to evaluate the PainSeeker model. Due to the limited number of samples with severe pain labels, we merge the moderate and severe samples into the "pain" category, allowing us to perform binary classification tasks. We used the leave-one-rat-out (LORO) protocol for evaluation, conducting $S$ folds of experiments, where $S$ represents the number of rats involved in the dataset. For each fold of the experiment, we use the facial images of one rat as the testing set, while the remaining rats' facial images are used as the training set.

The performance metrics employed are the \textit{F1-score} and \textit{Accuracy}, which can be calculated using the following formulas: \textit{F1-score} $=\frac{2TP}{2TP+FP+FN}$ and \textit{Accuracy} $=\frac{TP + TN}{TP + FP + FN + TN}\times 100$. Here, TP, FP, FN, and TN represent the numbers of the rats' facial images across all the folds correctly predicted as pain, incorrectly predicted as pain, incorrectly predicted as no pain, and correctly predicted as no pain labels, respectively. During the experiments, we manually cropped each rat's facial image from the original image, containing the four key facial components, as shown in Fig.~\ref{fig:nb1}(d). These cropped images are then resized to $224\times224$ pixels.

For PainSeeker, we utilize the convolutional layers of ResNet-18~\cite{he2016deep} to extract raw features from facial local regions and fix its trade-off parameter $\lambda$, margin value $\delta$ and number of highly pain-related facial local regions $K_{h}$ at 0.1, 0.05, and 5. During the training, the batch size, optimizer, and learning rate are set as 64, Adam optimizer, and $1e^{-4}$, respectively. Moreover, we also include the results obtained by the original ResNet-18 model for comparison. Additionally, we conduct experiments using several conventional machine learning methods, namely, the combination of local binary pattern (LBP)~\cite{ojala2002multiresolution} with different spatial divisions ($4\times4$, $8\times8$, and $16\times16$) and $R=\{1,3\}/P=8$ and linear support vector machine (SVM)~\cite{cortes1995support} with penalty coefficient $C = 1$.


\subsection{Results and Discussions}

The experimental results for all methods are presented in Table~\ref{tab:nb2}. From the table, it is evident that all methods can achieve promising performance in terms of both \textit{F1-score} and \textit{Accuracy},  when addressing the task of pain assessment in rats through facial expressions. They significantly outperform random guessing, providing experimental evidence to support the notion that pain in rats can be assessed entirely via their facial expressions. Additionally, it is worth noting that our PainSeeker method performs best among all comparison methods, achieving an impressive \textit{F1-score} of $0.7757$ and \textit{Accuracy} of $73.37\%$. This demonstrates the effectiveness and superior performance of the PainSeeker model in coping with this emerging and interesting topic.

Moreover, we conduct an ablation analysis for PainSeeker to examine the impact of the PRSC regularization on the model's performance. In this regard, we remove the PRSC term from the original PainSeeker model, resulting in a reduced version denoted as PainSeeker w/o PRSC. We then perform LORO experiments to compare the performance of this reduced version (with an \textit{F1-score} of 0.7513 and \textit{Accuracy} of 70.56\%) with the original PainSeeker model (with an \textit{F1-score} of 0.7757 and \textit{Accuracy} of 73.37\%). This comparison clearly shows that the inclusion of the PRSC loss effectively enhances the performance of PainSeeker for assessing pain in rats through facial expressions.

In addition, we also provide the \textit{F1-score} and \textit{Accuracy} for each rat achieved by all comparison methods in Table~\ref{tab:nb2}. The results reveal an interesting trend, as there consistently exists a significant gap between the performance of Rats\#3/4 and the other rats across almost all the methods. This observation prompts us to consider the possibility of individual differences among subjects in the tasks of pain assessment in rats via facial expressions, which can affect the effectiveness of both deep learning and conventional machine learning approaches. Further investigation is required to explore this aspect and develop more effective methods that can address this challenge, while also being pain-discriminative and pose-invariant among rats.



\section{Conclusion}

In this letter, we have addressed the issue of assessing pain in rats through their facial expressions by introducing the RatsPain dataset and proposing a deep learning method called PainSeeker. Our contribution lies in two aspects. Firstly, RatsPain is the first publicly available and well-labeled facial image dataset specifically designed for investigating pain assessment in rats through facial expressions. Secondly, the proposed PainSeeker tackles the common challenge of head pose variance in facial images of rats captured by fixed-view cameras. We conducted extensive experiments on the collected RatsPain dataset to evaluate the PainSeeker. The results not only demonstrate the effectiveness of the proposed method but also provide experimental evidence supporting the feasibility of pain assessment in rats through facial expressions.


%

%
%
%
%
%

\ifCLASSOPTIONcaptionsoff
\newpage
\fi



\bibliographystyle{IEEEtran}
\bibliography{SPL2023}
\end{document}